\documentclass[runningheads]{llncs}

\usepackage{graphicx}
\usepackage{hyperref}
\usepackage{xcolor}

\usepackage{booktabs}

\usepackage{amsmath}

\begin{document}
\title{Image-text matching for large-scale book collections}

%
%
\author{Artemis Llabrés\orcidID{0000-0002-6128-1796}\and
Arka Ujjal Dey\orcidID{0000-0001-8392-1574}\and
Dimosthenis Karatzas\orcidID{0000-0001-8762-4454}\and
Ernest Valveny\orcidID{0000-0002-0368-9697}
}
\authorrunning{A. Llabrés et al.}

\institute{Computer Vision Center, UAB, Spain\\
\email{\{allabres, audey, dimos, ernest\}@cvc.uab.cat}}

\maketitle             
\begin{abstract}
We address the problem of detecting and mapping all books in a collection of images to entries in a given book catalogue. Instead of performing independent retrieval for each book detected, we treat the image-text mapping problem as a many-to-many matching process, looking for the best overall match between the two sets.
We combine a state-of-the-art segmentation method (SAM) to detect book spines and extract book information using a commercial OCR. We then propose a two-stage approach for text-image matching, where CLIP embeddings are used first for fast matching, followed by a second slower stage to refine the matching, employing either the Hungarian Algorithm or a BERT-based model trained to cope with noisy OCR input and partial text matches.
To evaluate our approach, we publish a new dataset of annotated bookshelf images that covers the whole book collection of a public library in Spain. In addition, we provide two target lists of book metadata, a closed-set of 15k book titles that corresponds to the known library inventory, and an open-set of 2.3M book titles to simulate an open-world scenario.
We report results on two settings, on one hand on a matching-only task, where the book segments and OCR is given and the objective is to perform many-to-many matching against the target lists, and a combined detection and matching task, where books must be first detected and recognised before they are matched to the target list entries.
We show that both the Hungarian Matching and the proposed BERT-based model outperform a fuzzy string matching baseline, and we highlight inherent limitations of the matching algorithms as the target increases in size, and when either of the two sets (detected books or target book list) is incomplete.
The dataset and code are available at \url{https://github.com/llabres/library-dataset}

\end{abstract}
\section{Introduction}
\label{sec:introduction}

There are $2.8M$ public libraries in the world. Only in 2023 they managed $8,059M$ book loans\footnote{\url{https://librarymap.ifla.org/map}}. Tracking the assets of these book collections involves continuous inventory taking. Most libraries make use of RFID tags to speed up the process, however these systems tend to fail with high book density \cite{9796711} and are thus only used at check-out and return points. Knowing in real time the location of each book in the library can enable new services, and better tracking of assets. But creating a full inventory of the books on the shelves is still a tedious and time-consuming manual task that takes place a few times per year.

On a separate take, understanding what books appear on someone’s personal collection can tell us a lot about individuals. There has been a renewed interest in understanding \textit{shelfies} (photographs of one’s book shelf) during the pandemic \cite{allsop2022shelfies,allsop2022framing,dezuanni2022selfies,fletcher2019loveyourshelfie}. Contrary to the constrained inventory taking problem, this involves matching all instances of books in an image collection to an ``open'' list of possible titles.

In this work, we explore the automated inventory taking of books on shelves.

\begin{figure}[h]
\centering
\includegraphics[width=1.0\textwidth]{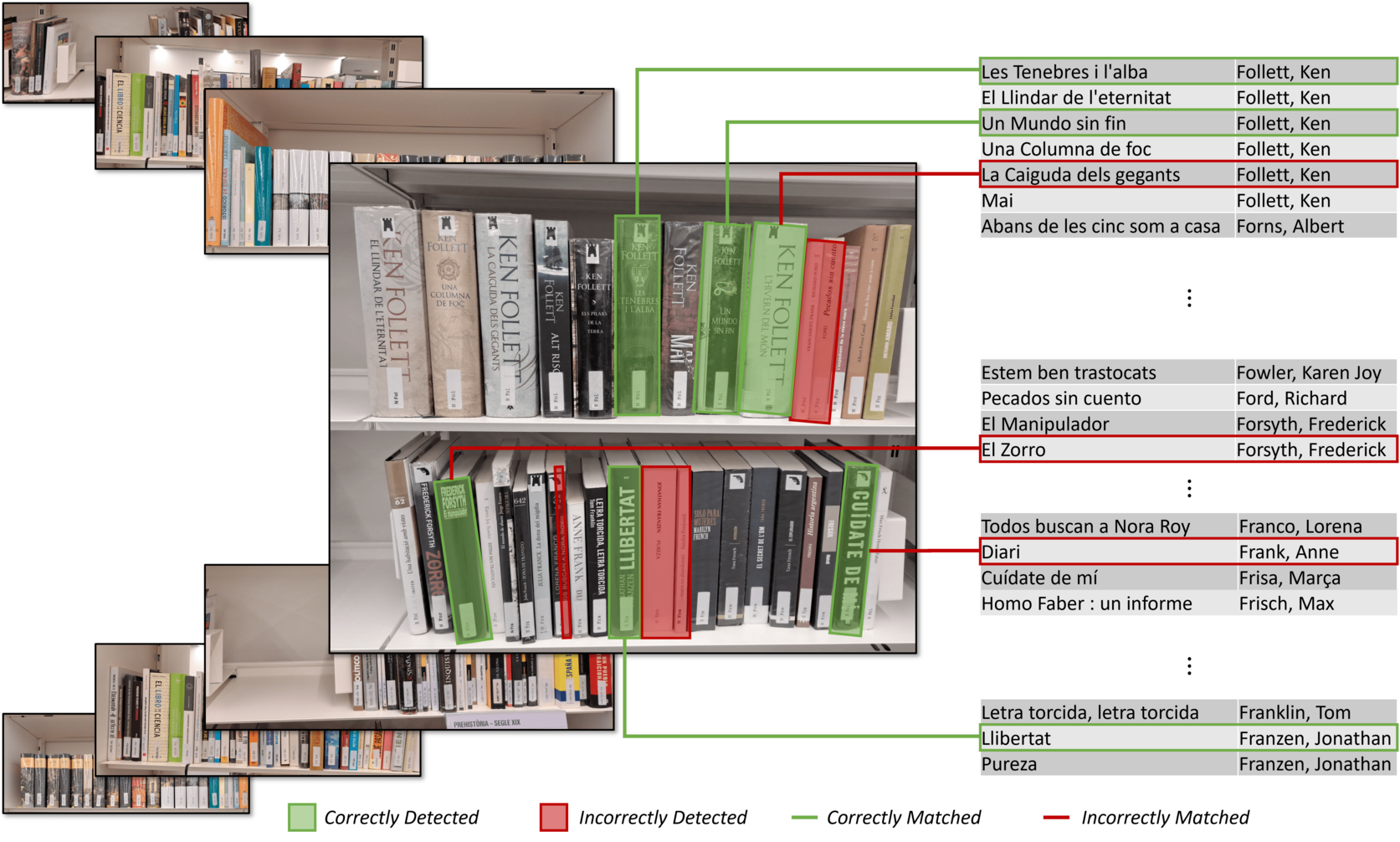}
\caption{The proposed task: books must be detected in a collection of images and subsequently matched in a many-to-many fashion against a target list.}
\label{fig:intro}
\end{figure}

Existing approaches for bookshelf analysis employ basic edge detection and Hough Transform operations \cite{7991581,7890886,10.1145/3277104.3277115} or image segmentation using deep learning \cite{10.1007/978-3-030-84760-9_27,10.1007/978-3-031-20309-1_36} to detect the book spines in the image. Then the OCRed text is typically used to retrieve the closest entry in a book dataset using some edit distance operation.

Such approaches perform poorly in real-life scenarios, where the book text is partially printed or the spine is partially visible due to occlusions. Although the book matching problem is inherently a many-to-many one, most existing approaches treat book matching as a retrieval problem, which is suboptimal.

The many-to-many matching problem consists in finding the best overall match between two sets. This is more complex than performing independent retrieval for each object, 
and can become intractable when the collections grow substantially in size.
In our work, we treat inventory taking as a many-to-many matching problem, where books appearing in multiple bookshelf images must be detected and matched to a target list of book metadata. We combine state of the art segmentation methods (SAM) and a commercial OCR system to detect book spines and extract information from the images. Then, we explore different approaches for this matching problem that are better suited for different scenarios: a true many-to-many matching using Hungarian matching over CLIP scores, and a BERT-based model finetuned to deal with noisy text inputs and partial text matches.

To explore this problem, we curate a new dataset captured in a public library in Barcelona, Spain, and define two scenarios: a closed-set and an open-set one. The closed-set scenario reflects the real-life problem of inventory taking, where the target list of book titles corresponds to the known library collection, and all books that appear in the images are expected to match to an entry in this target list. In the closed-set scenario, the library’s catalogue is the target collection. The open-set scenario aims to map all books in a set of images to a much large “open-set” list of books, that include $2.3M$ titles. This is a much more demanding scenario, that would correspond to the analysis of \textit{shelfies} in uncontrolled settings.

Specifically, our contributions are the following:
\begin{itemize}
    \item We provide a large, annotated dataset of images captured at a public library in Barcelona, Spain that comprises titles in multiple languages. The dataset comprises $7,536$ books on $285$ bookshelf images. We also provide two book catalogues for matching, the true library inventory of $15,229$ entries (closed-set scenario) and a large-scale catalogue of $2.3M$ books (open-set scenario).
    \item We provide annotations of all titles visible in every image of the dataset, as well as additional weak annotations in the form of manually filtered SAM-produced segmentation maps and commercial OCR results for each image.
    \item We define two tasks: first, a “matching-only” task, where book segmentations and noisy OCR information are provided, and the problem is to match the detected and transcribed books to the target list. Second, a combined “detection and matching” task where books must first be detected and recognised and subsequently matched to entries in the target list.
    \item We report results of book matching both in the closed-set and the open-set scenario. We show that Hungarian Matching is best when lists are small but does not scale up well. We highlight inherent limitations of the algorithm when either of the two sets (detected books or target book list) is incomplete.
    \item We explore a BERT-based model trained to cope with noisy and incomplete text matching, as an alternative matching method. Although this model yields worse performance, it copes better with incomplete lists.
\end{itemize}

\section{Related Work}
\subsection{Text-Image Retrieval}

The task of text-image retrieval consists of either finding the most similar text caption for a given image or finding the most similar image given a caption. In current methods, such as CLIP \cite{radford2021learning}, this is achieved by creating a joint representation of text and image embeddings where image-text similarity can be easily computed. However, when working with two sets, these are not matched as a many to many problem, where more than one object of the first set can be matched to the same object on the second. This is why, in our approach, we introduce second stage using the Hungarian Algorithm.

\subsection{Automatic Book Inventory}

 The creation of an automatic book inventory requires the detection of book spines in bookshelve images. In most existing approaches this is done using edge detectors and Hough transforms to get the vertical lines that separate the books \cite{10.1007/978-3-030-84760-9_27,10.1007/978-3-031-20309-1_36}. If the images contain more than one shelf, first these are separated by detecting horizontal lines \cite{7890886}. Then the text on each book spine is read using Optical Character Recognition (OCR).

Approaches like \cite{10.1007/978-3-031-20309-1_36} create the book inventory by simply adding the text read from each book spine to a database. However, this means that errors done during reading will be in the inventory, hence why most approaches have a last matching stage, where the text read from the book spines is matched to a list of books.

In \cite{10.1145/3277104.3277115} only the tags used in libraries to organise books are read and used for the matching. This results in a much simpler reading problem, as all tags have the same font on a white background. Despite the books on our images having this kind of tags, we do not use them, we do the matching between the full spine text of the books and a list that contains the title and authors of books. This way our approach is more general and will work with any book and in different scenarios.

\begin{figure}[h!]
\centering
\includegraphics[width=0.7\textwidth]{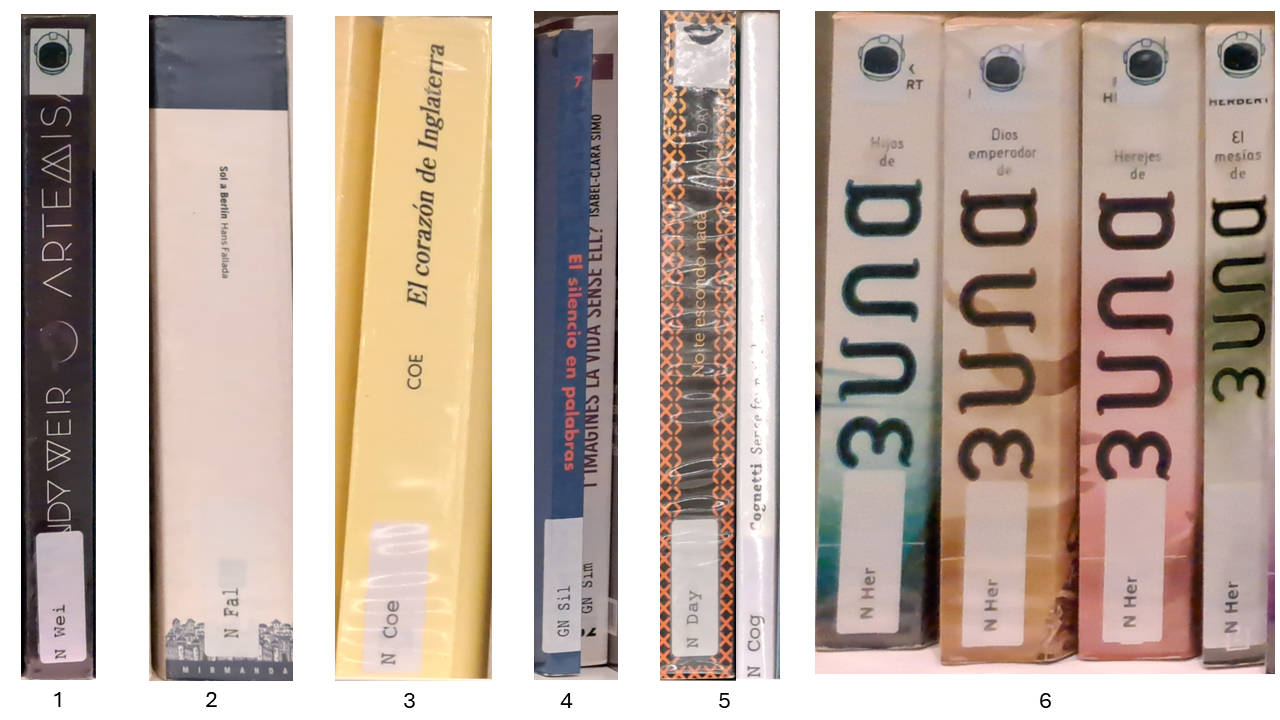}
\caption{Example book spines. \textit{1.} Weird font. \textit{2.} Small font. \textit{3.} Only author name present. \textit{4.} One book occluding the other. \textit{5.} Reflections making the text hard to read. \textit{6.} Series name larger than book title.}
\label{fig:book_spines}
\end{figure}

Matching the text in a book spine to an author and title is not as easy as it may seem. Book spines are narrow and offer little space, and hence author names are sometimes abbreviated, or if there are multiple authors, only one is present in the spine, or if the book has a subtitle, only the title is present, or if the book belongs to a series, the series title might be bigger than the book title, and many other cases which will make the text on the book spine very different from the text that could be on a book database used as target list. This is further complicated by the fact that the OCR will not perfectly read the text which not only tends to be on a small font on a very large image, but also occlusions and reflections might render some parts illegible. Some examples can be seen in Figure \ref{fig:book_spines}.

In \cite{10.1007/978-3-030-84760-9_27} the matching is done using fuzzy string matching between the OCR text and a target list of book titles. The matching is not done many to many, and is not large-scale as only 50 bookshelve images are used.

Another approach that does not require an explicit OCR is to match the image features extracted with a CNN to the image features on a database\cite{ZHOU2022103101}, the problem of this approach is that it requires to setup a database of pictures of the book spines beforehand, which is of course time-consuming. That is why we think that the OCR methods that match the text of the book spine to an entry on a target list of book author+title are better, as they are simpler to setup.

However, in all this cases, the matching is approached as a retrieval problem of finding for each book the most similar target, instead of considering it as a many to many matching task, where each matching is not independent, but rather done simultaneously.

\section{The Library Dataset}
\label{sec:dataset}
\subsection{Dataset Description}
\begin{figure}[h!]
\includegraphics[width=0.5\textwidth]{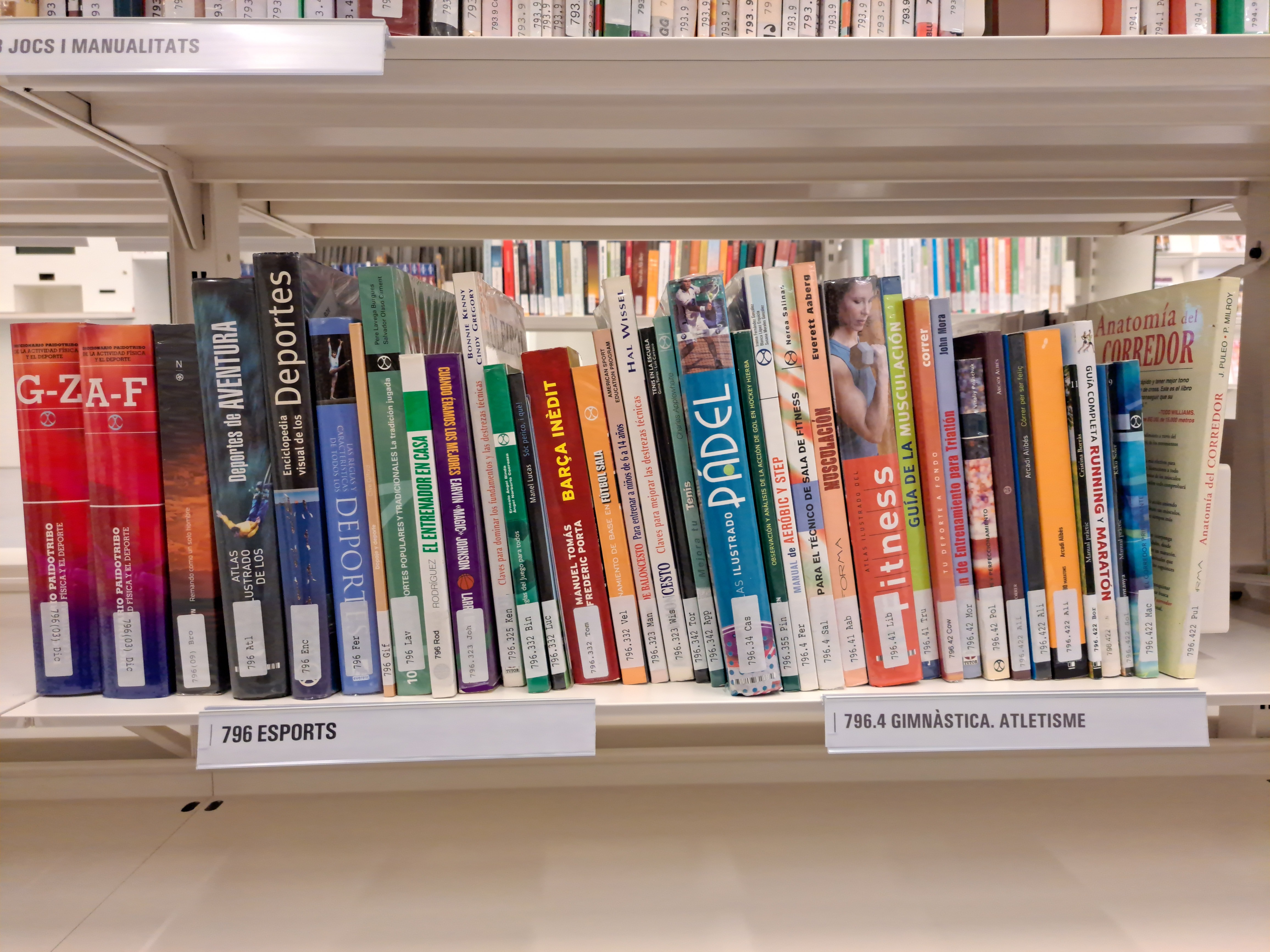}
\includegraphics[width=0.5\textwidth]{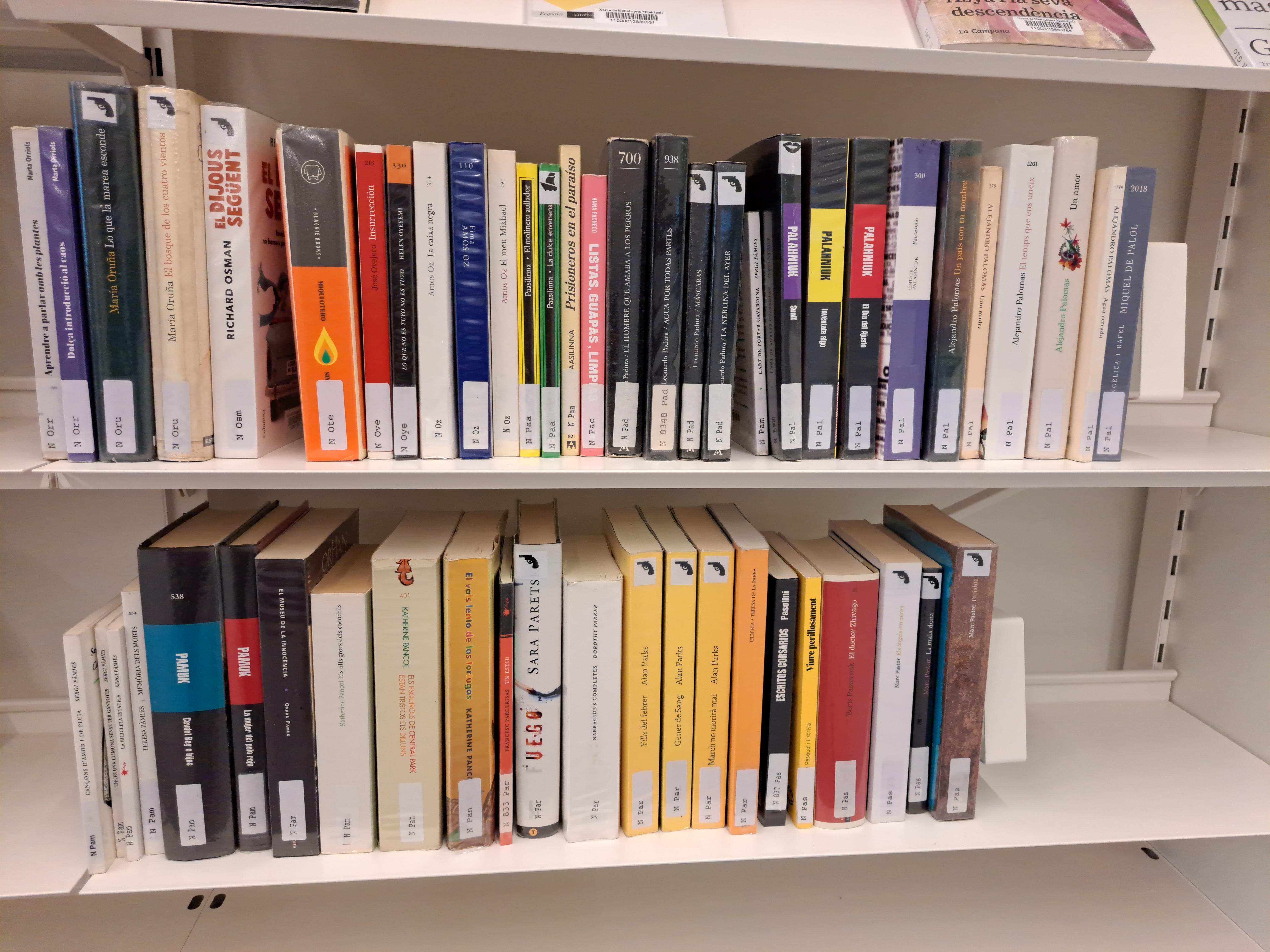}
\caption{\textit{left} Example image with one shelf. \textit{right} Example image with two shelves}
\label{fig:example_images}
\end{figure}

The Library Dataset consists of 285 high resolution images of bookshelves, like the ones in Fig.~\ref{fig:example_images}, taken at a public library. Each image contains either one or two shelves, and in some cases the top or bottom of the books on the shelf above or below are visible which, as we discuss in \ref{sec:hard},  will make the task more realistic, but also harder.

\begin{figure}[h!]
\centering
\includegraphics[width=0.7\textwidth]{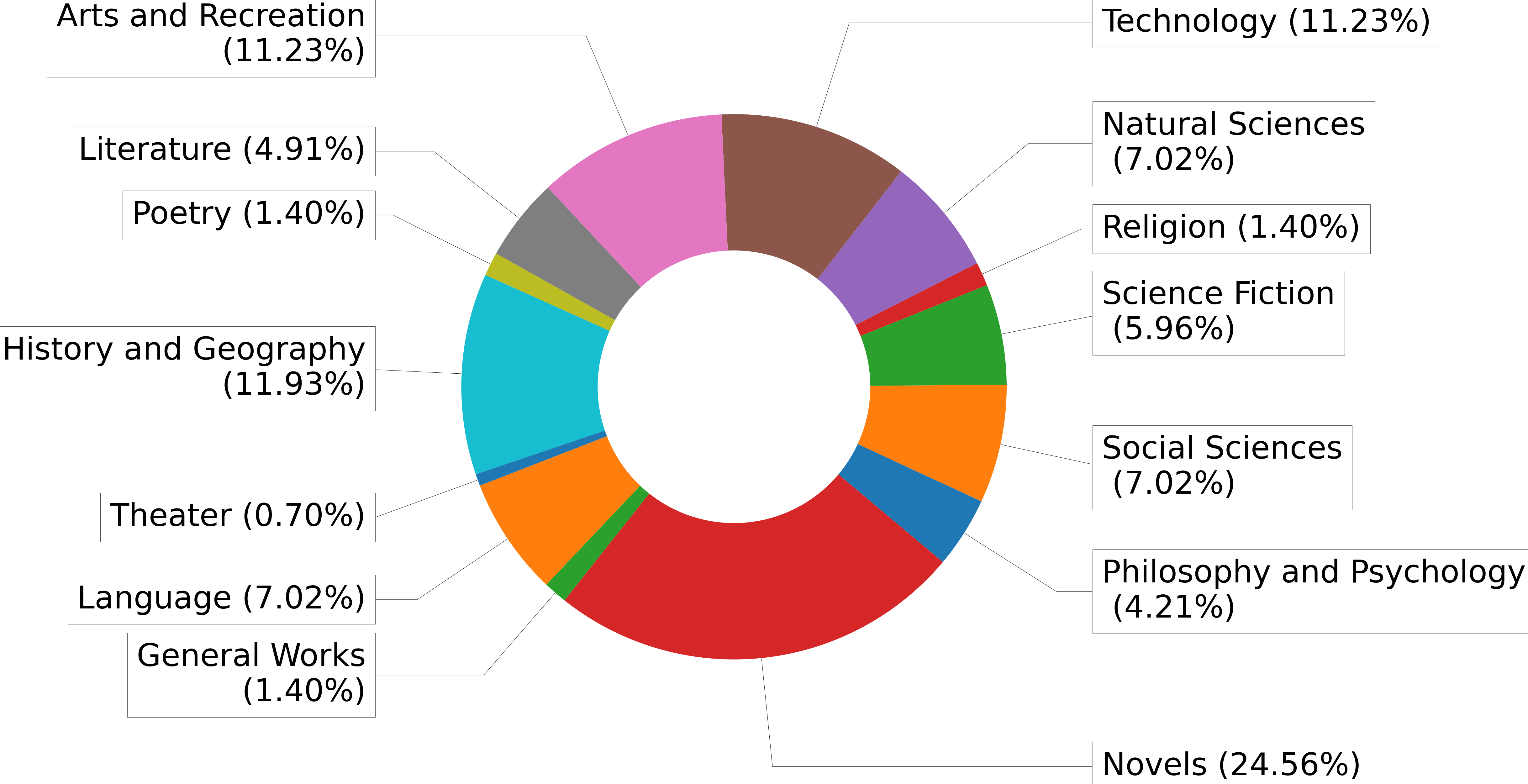}
\caption{Pie Chart with the number of images per each section of the library.}
\label{fig:library_sections}
\end{figure}

There are in total 7536 books in the images. In the large majority of cases only the book spine can be seen, with a few exceptions where the book cover is visible. There is a high diversity in the type of books as the images are from 14 different sections of the library, as can be seen in Figure \ref{fig:library_sections}. Also the books are in three main languages: Catalan, Spanish, and English, in that order, with some other languages in much smaller proportions: French, German, Italian, and Arabic.

In order to treat this as a many to many matching problem a target list is needed. We provided two book lists with the author, title, and ISBN of each book:
\begin{itemize}
    \item Library List: which contains 15229 books. This list was provided by the library and hence contains almost every book that can be seen in the images.
    \item Large list of Popular Books: which contains 2.3 million books from the website GoodReads, including the ones from the library list. This list is used to test the benchmarks at scale.
\end{itemize}

As mentioned the target lists contain almost every book that is in the library images, but some books are missing. This will mean that some books in the images will never be correctly matched with a book from the list. While this may seem unfair to the baseline methods, this makes for a more realistic scenario. In the annotations these books are set as `not in list', and hence more advanced methods should leave them unmatched.

Therefore, the dataset consists of a set of 285 images, two target lists, and the ground truth annotations of which books appear in each image.

\subsection{Data Collection and Annotation}
\label{sec:annotation}
In order to annotate the dataset a manual inventory of the library would be needed which would be very time-consuming. However, since the baselines described in section \ref{sec:methods} do not require supervised training, we can run them to generate a first automatic annotation that can then be manually checked and refined.

To generate the automatic annotation we first used Amazon's Rekognition OCR which gets us both the text that appears in the images and its location, and second we used Segment Anything (SAM)~\cite{kirillov2023segment} to automatically get masks for every object in the images. After postprocessing, masks that contain text detected by the OCR are considered useful and the rest are discarded.

\begin{figure}[h!]
\includegraphics[width=\textwidth]{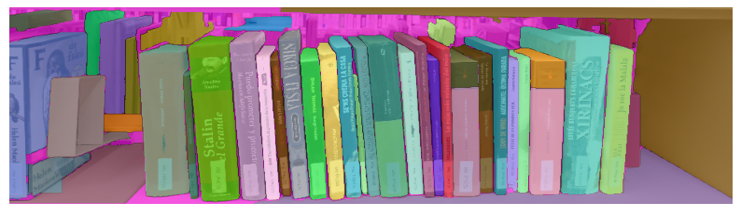}
\caption{Example of SAM's segmentation after postprocessing, and only keeping objects that contain text. While mostly correct, there are errors with two-coloured book spines segmented as two different objects, or two books that get merged together.}
\label{fig:book_segmentation}
\end{figure}

As can be seen in Fig.~\ref{fig:book_segmentation} this process is not perfect, as some objects that have text, like the section labels are segmented, or books that have two-coloured spines are segmented into two, or multiple books are segmented as one, or partially visible books from other shelves are also identified as a book.

These potentially detected books are manually checked, and classified as `book' if they are a correctly segmented full book, `not a book' if they are not a book or a partially segmented book, and `merged books' if more than one book is segmented together. The results of this manual classification can be seen in Table \ref{tab:segmentation_results}.

\begin{table}
\centering
\caption{Number of segmented objects by SAM as `Total Objects', number of segmented objects that contain text as `Objects with text', and the results of manually classifying these objects as `book', `not a book', and `merged books'.}\label{tab:segmentation_results}
\begin{tabular}{|c|c|c|c|c|}
\hline
 Total objects  & Objects with text  & Books & Not books & Merged books\\
\hline
9850 & 8435 & 6912 & 1267 & 256\\
\hline
\end{tabular}
\end{table}

Using the segmented objects manually classified as books the CLIP + Hungarian approach explained at section \ref{sec:methods} is used to match each book with a book from the library list. Then these matches are manually checked and corrected if necessary.

Finally to complete the annotations, the books that were not correctly segmented of each image are manually annotated.

Therefore the dataset consists of 285 images of bookshelves containing a total 7536, with annotations of which books are in each image, plus two target lists of 15229 and 2.3 million books respectively. Also, despite not being perfect, both the OCR and SAM's segmentation are provided as part of the dataset.

\section{Baselines and Methods}
\label{sec:methods}
The baseline and methods proposed only tackle the matching problem, since they work on top of the OCR + segmentation pipeline explained on the annotation section \ref{sec:annotation}. 

\subsection{Fuzzy String Matching Baseline}
This is a simple baseline that uses approximate string matching or fuzzy string matching~\cite{max_bachmann_2021_5584996}, to compute the similarity between the text on the book spines read by the OCR with the author + title that are on the target list.

The similarity is computed as follows:
$$\textit{similarity}= 1 - \textit{normalized\_distance} = 1 - \frac{\textit{distance}}{\textit{max}}$$
Where \textit{distance} is the Levenshtein distance \cite{Levenshtein1965BinaryCC} between OCR text and target, and \textit{max} is the maximal possible Levenshtein distance given the lengths of the two strings.

Therefore each book will be matched with the target book with the largest similarity.

While being the most often used approach for book matching, it is not the most convenient, since text in the spine is often quite different from the actual author+title found in the target list. The other problem with this approach is its speed, specially when working at large scale as the amount of comparisons needed to compute all the similarities drastically increases.

\subsection{A two stage approach for many to many matching}
\begin{figure}[h!]
\centering
\includegraphics[width=\textwidth]{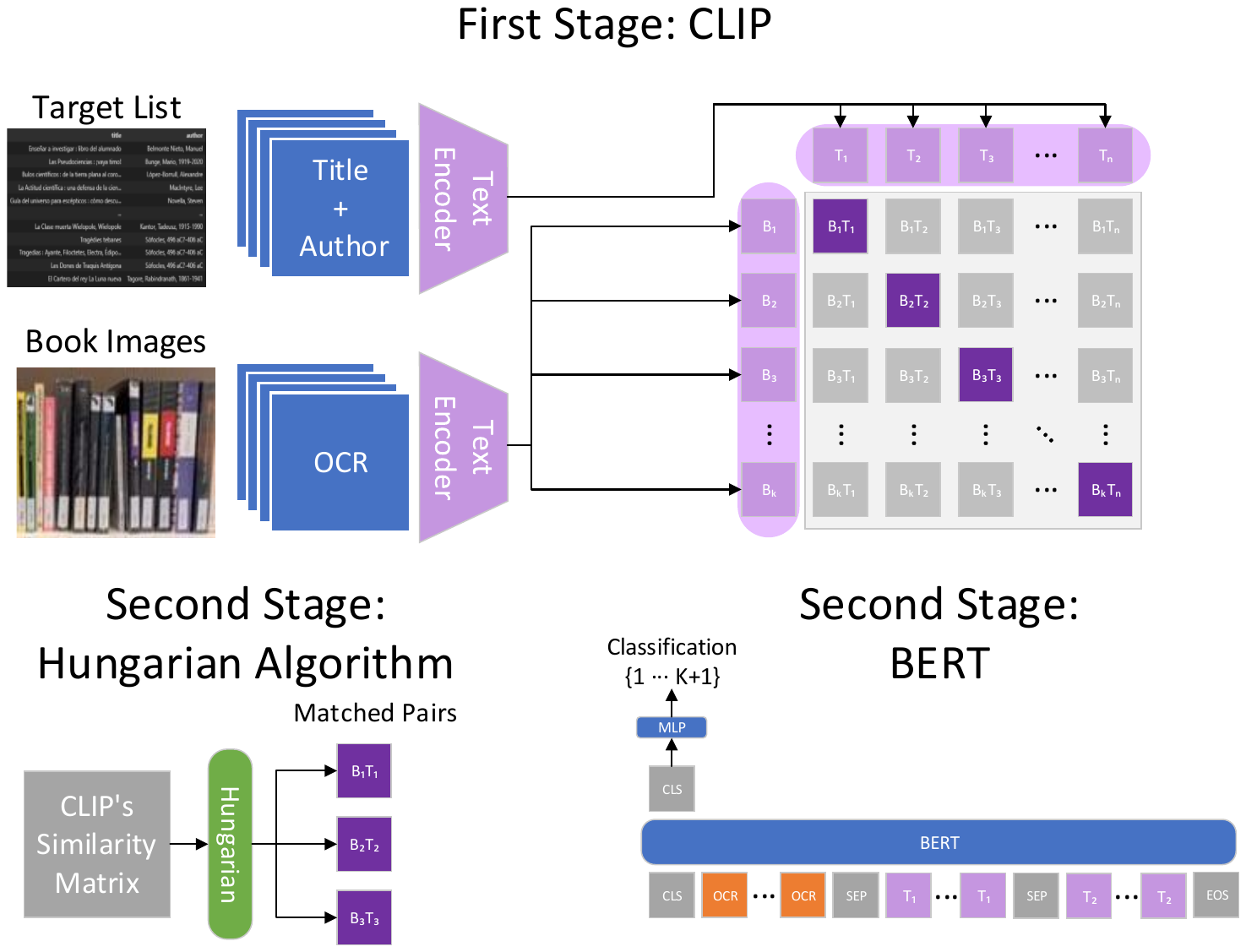}
\caption{Two stage approach using CLIP as a first stage, and two possible second stages Hungarian Algorithm, and BERT.}
\label{fig:pipeline}
\end{figure}

Following inspiration from \cite{miech2021thinking}, we propose a two stage method (see Figure \ref{fig:pipeline}), that uses CLIP as a fast first stage to compute a similarity matrix between all books and all targets, and then a second stage, which is slower, to refine the matching. We propose two possible alternatives for the second stage, one using the Hungarian Algorithm, and another one using BERT.

\subsubsection{First Stage: CLIP}
Using CLIP's~\cite{radford2021learning} text encoder we embed the OCR text of every book that has been segmented in the images, which gives us a matrix $B$ (B of books) with shape $N_B\times D$, where $N_B$ is the total number of books and $D$ the embedding size. We also embed the author+title of every book in the target list, which gives us a matrix $T$ (T of targets) with shape $N_T\times D$, where $N_T$ is the total number of target books

By computing the matrix multiplication $BT^T$ we obtain a similarity matrix $S$ with shape $N_B \times N_T$, with at each entry the cosine similarity between each row of $B$, meaning each book, and each column of $T^T$ meaning each target.

Each book is matched with the target with highest similarity, this means: that the $i$ book will be matched with the $j=\text{argmax}(S_i)$ target, where $S_i$ is the i-th row of the matrix $S$.

\begin{figure}[h!]
\centering
\includegraphics[width=0.7\textwidth]{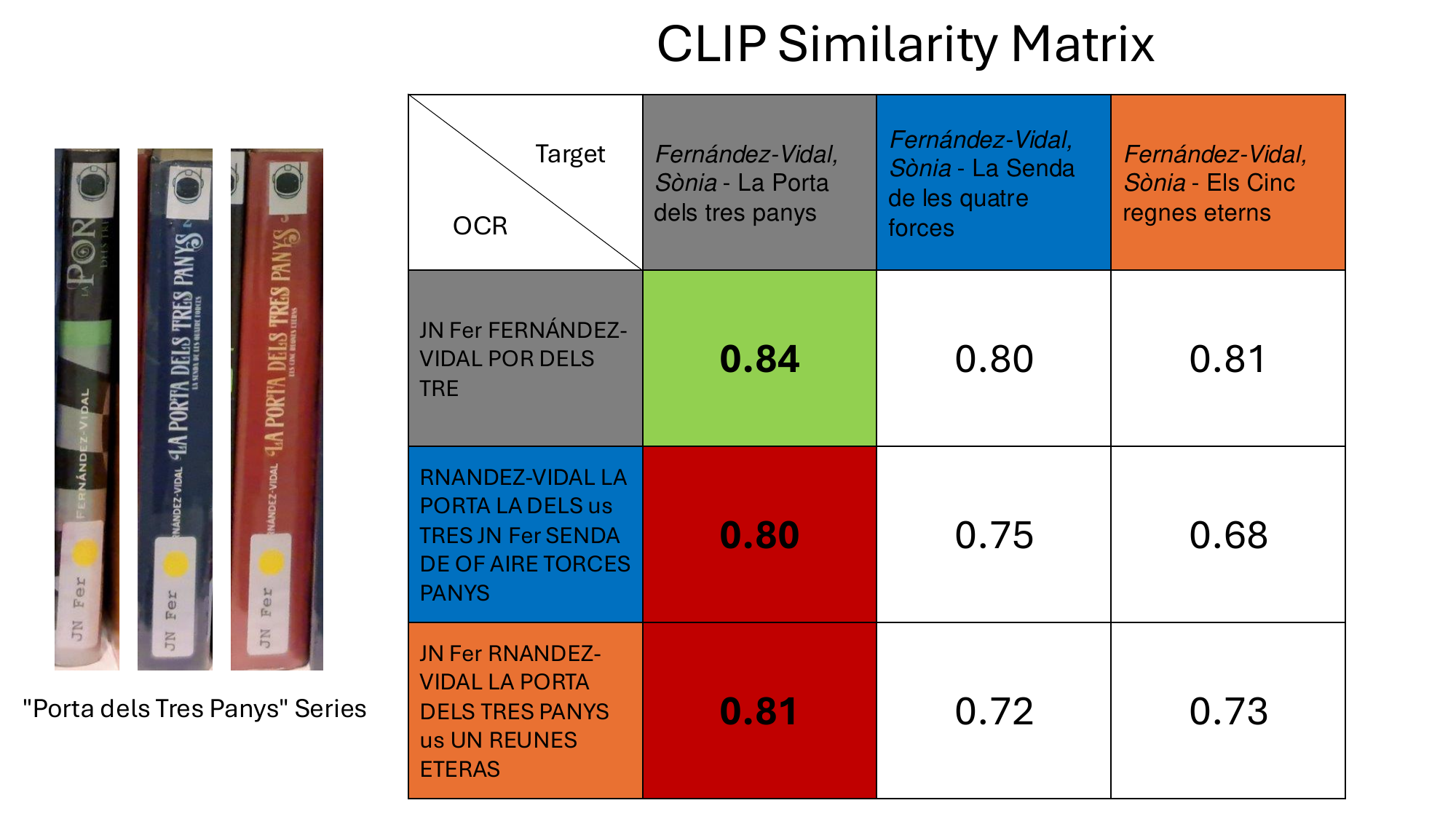}
\caption{Example of matching failure, since the OCR of the three books contains the author name, and the name of the series, which is also the name of the first book, the highest similarity of all three books is with the target of the first book, instead of the correct one for each.}
\label{fig:example_failure}
\end{figure}

The main problem with this approach is that the matching is done independently for each book, selecting only the target with highest similarity. At no point collisions with other books are taken under consideration, meaning that two books could end up matched to the same target. In Figure~\ref{fig:example_failure} we see an example where three books that belong to the same series are incorrectly matched to the same target. This happens because the series name is also the name of the first book. If we only allowed one book to be matched with one target, then these books would all be correctly matched, showing the need for a second stage to refine the matching.

\subsubsection{Second Stage: Hungarian Algorithm}
In order to refine CLIP's matching and solve the issue of two different books being matched to the same target we propose to use the Hungarian Algorithm~\cite{https://doi.org/10.1002/nav.3800020109}, which solves the assignment problem over CLIP's Similarity Matrix $S$, by finding the matching of one-to-one pairs of books and targets that has overall the highest similarity.

While this approach refines the matching improving results it comes at the cost of speed, making it non-viable for the largest 2.3 million book target list.

\subsubsection{Second Stage: BERT}
One issue that neither CLIP nor the Hungarian Algorithm is tackling is the possibility of books not appearing in the target list or that the text-object that has been segmented is not a book, since they will always match every object or book to a book on the target list.

\begin{figure}[h!]
\includegraphics[width=\textwidth]{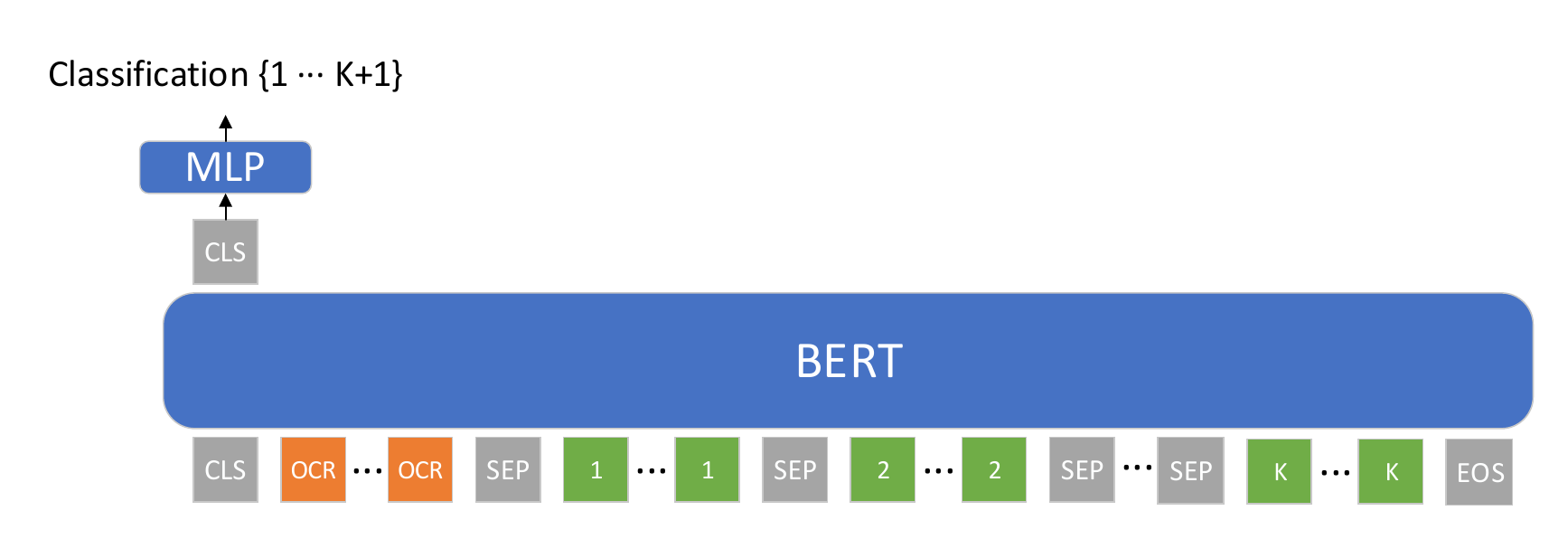}
\caption{After CLIP has selected the top $K$ most similar matches, the tokens from the OCR text (in orange), and the tokens for the $1$ to $K$ matches (in green) are embedded by BERT. On the CLS (in grey) token classification is performed selecting the correct target or the extra id if none of the $K$ targets are a good match.}
\label{fig:bert}
\end{figure}

To solve this issue we introduce a different option for a second stage based on BERT~\cite{devlin2019bert}. The idea is to use CLIP's similarity matrix between books and targets as a filter to only keep the top $K$ most promising matches for each book. Then these top $K$ predictions (where $K$ is much smaller than the size of the target list), can be processed by BERT as can be seen in Figure \ref{fig:bert}, embedding each book together with its $K$ possible targets. The CLS token in the output is passed through a linear layer and a softmax layer to predict the index of the best target match or an extra special index meaning that none of the targets are a good match for this book.

Despite starting from BERT's pretrained weights, for this approach to work, some task-specific fine-tuning is required. To do so, we defined a specific fine-tuning task where the goal is to match corrupted versions of the books of the list, to the pristine versions. For that, we created a synthetic dataset from the book target list, where corruptions consist of deleting or replacing letters or even deleting and replacing entire words. In this way we simulate the differences that there are between the text on book spines and the text (title and author) found on the target list, and also possible OCR reading errors.

The fine-tuning is done following an approach similar to \cite{li2021align}. We randomly select a book from the target list and create a corrupted version which will simulate the book-spine text read by the OCR.
We use a frozen CLIP to find the $K$ most similar samples from the target list. 50\% of the time we include the correct book in the $K$ samples, and 50\% we do not. Then, we use BERT to embed the corrupted book together with the $K$ targets. If we have included the correct target in the $K$ samples, BERT should predict its index in the CLS token. Otherwise, it should predict the extra index that means that the book is not matched with any of the samples.

For fine-tuning and testing in the experiments below, $K$ is set to 10.

\section{Experiments}
\label{sec:experiments}
\subsection{Matching Only task}
\label{sec:easy}
On this simplified problem the inputs are only the spines of the books that were correctly segmented instead of the full images, hence only the matching ability of the methods is being evaluated. The task is further simplified by only including books that appear on the target lists, meaning that every book spine will have a correct match in the list.

In Table \ref{tab:easy_results}, the baseline and our methods are tested when matching the books against three different lists: \textit{exact}, \textit{library}, and \textit{all}.

\textit{Exact} is a list that contains exactly the same books that are in the images. Therefore, there is a perfect one-to-one matching between the books in the images and the target list.

\textit{Library} is the list that was provided by the library, which contains 15229 books. Therefore, there will be more target books than books in the images, we call these extra books distractors.

\textit{All} is the list that contains the most popular books, 2.3 million in total. Therefore, the number of target books is overwhelmingly larger than the number of books in the images, meaning that there will be  many distractors that will difficult the task of finding the correct matches. 

The best performance, in the \textit{exact} list is obtained by the two stage method using CLIP and the Hungarian Algorithm. While on \textit{library}, this method still obtains the best accuracy, it is only 0.003 over the accuracy obtained by CLIP on its own. This showcases the problem with the Hungarian Algorithm, its performance suffers as the number of distractors (number of books that are in the target list, but not in the images) increases. Another problem with the Hungarian Algorithm is its time and memory requirements making it computationally unfeasible for the largest target list.

Performance drastically drops when using the \textit{all} list, showing that even in this simplified problem there is still work to do to obtain fast many to many matching methods with high accuracy.

\begin{table}[h!]
    \centering
    \caption{Accuracy using all books and target lists: exact, library, and all}
\begin{tabular}{|l|c|c|c|}
\hline
methods &   exact &  library &     all \\
\hline
String matching &  0.783 & 0.742 & \textbf{0.505}  \\
CLIP         &  0.939 &   0.915 &  0.449 \\
CLIP + hungarian      &  \textbf{0.984} &   \textbf{0.918} &       - \\
CLIP + BERT & 0.836 & 0.824 &  0.310 \\

\hline
\end{tabular}
    \label{tab:easy_results}
\end{table}

\subsection{Detection and Matching Task}
\label{sec:hard}
This task tackles the full problem of creating a book inventory. The input are the 285 bookshelves images, this means that unlike in the simplified task there can be books that appear in the images, but not on the target list, and therefore should be classified as `Not on the list'. Instead of only using the correctly segmented books, the baseline and methods will have to deal with the real output of the segmentation models, which contains errors like partially visible books from other shelves, books merged together by the segmentation, or other objects with text that are not books, like signs that indicate the library section.

The results of this experiment are on Table \ref{tab:hard_results}. There is a significant drop in performance from the previous experiment, but the trend stays the same, CLIP + hungarian achieves the highest accuracy on the \textit{library} target list.

\begin{table}[h!]
    \centering
    \caption{Accuracy using all books and target lists: library, and all}
\begin{tabular}{|l|c|c|}
\hline
methods   &  library &     all \\
\hline
String Matching  & 0.573 &  \textbf{0.389}  \\
CLIP             & 0.617 &   0.241  \\
CLIP + hungarian & \textbf{0.641} &   -  \\
CLIP + BERT      & 0.622 &  0.240 \\

\hline
\end{tabular}
    \label{tab:hard_results}
\end{table}

\section{Conclusions}
\label{sec:conclusions}
In this work we have proposed a two-stage method based on CLIP, for the first stage, and two variants, Hungarian Algorithm and BERT, for the second stage, to tackle many to many matching between text-objects in multiple images and a large offline text corpus. In particular we study the case of automatic inventory of books, by introducing a new dataset of bookshelves images and two target lists. We test our proposed method, which improves over the fuzzy string matching baseline when using the small target list and with the larger target list we show the need for further research in this area.

\section*{Acknowledgements}
We want to thank Laura Solà from the Library of Volpelleres Miquel Batllori for the help given during the collection of the dataset.

%
%

\bibliographystyle{splncs04}
\bibliography{bibliography}

\end{document}